\documentclass[journal, letterpaper, twoside, 10 pt]{IEEEtran}

\usepackage{ifpdf}
\usepackage{cite}

\ifCLASSINFOpdf
\usepackage[pdftex]{graphicx}
\graphicspath{{../pdf/}{./figures/}}
\else
\fi

\usepackage{amsmath}
\usepackage{url}
\usepackage{siunitx}
\usepackage{nicefrac} 
\usepackage{caption}
\usepackage{subfig}
\usepackage[colorlinks = true, urlcolor = blue,citecolor = black, linkcolor = black]{hyperref}
\usepackage{adjustbox}
\usepackage{balance}
\usepackage[font=footnotesize]{caption}
\usepackage{threeparttable}
\usepackage{soul}
\usepackage{verbatim}
\usepackage{amssymb}
\usepackage{mathtools}
\usepackage{xparse}

\makeatletter
\newcommand{\raisemath}[1]{\mathpalette{\raiseMath{#1}}}%
\newcommand{\raiseMath}[3]{\raisebox{#1}[0pt][0pt]{$#2#3$}}
\makeatother

\NewDocumentCommand{\qbar}{O{0.5pt} O{-6.55pt}}{
	\ensuremath{\mathrlap{\raisemath{#2}{\hspace*{#1}{\mathchar'26\mkern-9mu}}} q}%
}

\NewDocumentCommand{\pbar}{O{-1.5pt} O{-6.65pt}}{
	\ensuremath{\mathrlap{\raisemath{#2}{\hspace*{#1}{\mathchar'26\mkern-9mu}}} p}%
}

\newcommand{\bs}[1]{\boldsymbol{#1}}  
\newcommand{\ts}[1]{\text{#1}} 

\renewcommand{\baselinestretch}{0.952}  

\usepackage{tikz}
\newcommand\copyrighttext{%
	\footnotesize \copyright 2019 IEEE.  Personal use of this material is permitted.  Permission from IEEE must be obtained for all other uses, in any current or future media, including reprinting/republishing this material for advertising or promotional purposes, creating new collective works, for resale or redistribution to servers or lists, or reuse of any copyrighted component of this work in other works.}
\newcommand\copyrightnotice{%
	\begin{tikzpicture}[remember picture,overlay]
	\node[anchor=south,yshift=5pt] at (current page.south) {\fbox{\parbox{\dimexpr\textwidth-\fboxsep-\fboxrule\relax}{\copyrighttext}}};
	\end{tikzpicture}%
}

\begin{document}

\title{Bee\textsuperscript{+}: A 95-mg Four-Winged Insect-Scale Flying Robot Driven by Twinned Unimorph Actuators}

\author{Xiufeng Yang,
        Ying Chen,
        Longlong Chang,
        Ariel A. Calder\'on,
        and~N\'estor~O.~P\'erez-Arancibia
\thanks{Manuscript received: February 24, 2019; Revised: May 29, 2019; Accepted: June 17, 2019.}
\thanks{This paper was recommended for publication by Editor Yu Sun upon evaluation of the Associate Editor and Reviewers' comments.}
\thanks{This work was supported by the \textit{National Science Foundation} (NSF) through NRI Award\,1528110 and the USC Viterbi School of Engineering.}
\thanks{The authors are with the Department of Aerospace and Mechanical Engineering, University of Southern California (USC), Los Angeles, CA 90089-1453, USA (e-mail: {\tt xiufeng@usc.edu; chen061@usc.edu; longlonc@usc.edu; aacalder@usc.edu; perezara@usc.edu}).}
\thanks{Digital Object Identifier (DOI): see top of this page.}}

\markboth{IEEE ROBOTICS AND AUTOMATION LETTERS. PREPRINT VERSION. ACCEPTED JULY, 2019}%
{Yang \MakeLowercase{\textit{et al.}}: Bee\textsuperscript{+}: A 95-\MakeLowercase{mg} Four-Winged Insect-Scale Flying Robot Driven by Twinned Unimorph Actuators}

\maketitle

\copyrightnotice

\vspace{-2.0ex}
\begin{abstract}
We introduce Bee\textsuperscript{+}, a 95-mg four-winged microrobot with improved controllability and open-loop-response characteristics with respect to those exhibited by state-of-the-art two-winged microrobots with the same size and similar weight (i.e., the 75-mg Harvard RoboBee). The key innovation that made possible the development of Bee\textsuperscript{+} is the introduction of an extremely light (28-mg) pair of twinned unimorph actuators, which enabled the design of a new microrobotic mechanism that flaps four wings independently. A first main advantage of the proposed design, compared to those of two-winged flyers, is that by increasing the number of actuators from two to four, the number of direct control inputs increases from three to four when simple sinusoidal excitations are employed. A second advantage of Bee\textsuperscript{+} is that its four-wing configuration and flapping mode naturally damp the rotational disturbances that commonly affect the yaw degree of freedom of two-winged microrobots. In addition, the proposed design greatly reduces the complexity of the associated fabrication process compared to those of other microrobots, as the unimorph actuators are fairly easy to build. Lastly, we hypothesize that given the relatively low wing-loading affecting their flapping mechanisms, the life expectancy of Bee\textsuperscript{+}s must be considerably higher than those of the two-winged counterparts. The functionality and basic capabilities of the robot are demonstrated through a set of simple control experiments.
\end{abstract}


\IEEEpeerreviewmaketitle

\vspace{-0.5ex}
\section{Introduction}
\vspace{-0.5ex}
\label{SECTION01}
\IEEEPARstart{I}{nsect-sized} aerial robots have the potential to be employed in a great number of tasks such as infrastructure inspection, search and rescue after disasters, artificial pollination, reconnaissance, surveillance, et cetera, which has motivated the interest of many research groups. Consistently, as an emerging field, research on cm-scale flapping-wing robots driven by piezoelectric actuators has produced numerous design innovations over the course of more than two decades \cite{Wood2008,perez2011first,ma2013controlled,perez2015JINT}. However, state-of-the-art flying microrobots, such as those reported in~\cite{graule2016perching} and \cite{fuller2019four}, do not adequately replicate the astounding capabilities exhibited by flying insects. An obstacle that has limited progress is the fact that unlike insects which simultaneously use multiple distributed muscles for flapping and control \cite{dickinson1997function}, flapping-wing flying robots are driven by a small number of discrete actuators due to stringent constraints in size and weight as well as fabrication challenges. 
\begin{figure}[t!]
\vspace{1ex}
\begin{center}
\includegraphics[width=3.4in]{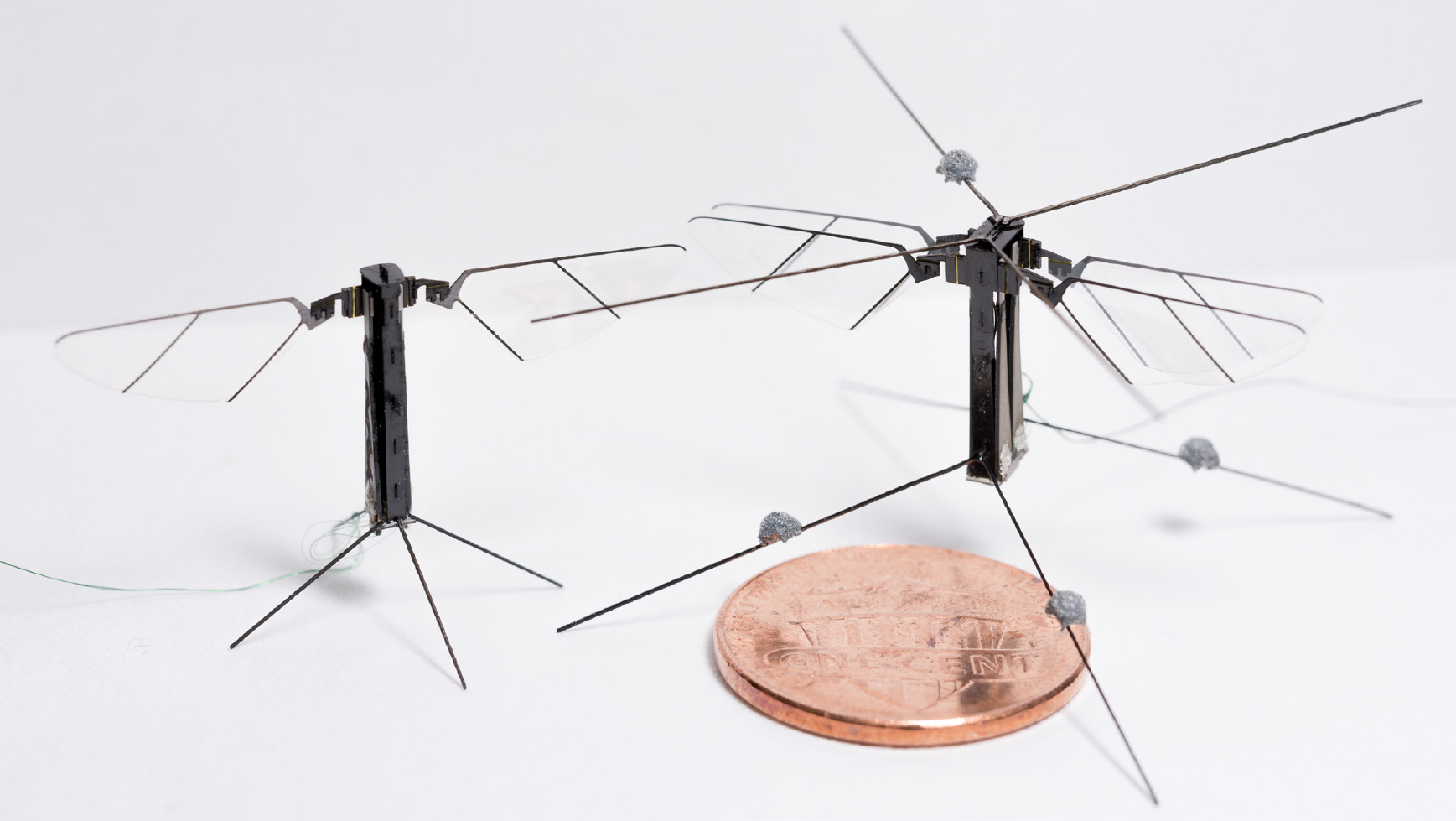}
\caption{Bee\textsuperscript{+} (right), a new four-winged flying microrobot. This robot has a mass of \SI{95}{\mg} and measures \SI{33}{\mm} in wingspan. Four retroreflective markers (\SI{5}{\mg}) for motion tracking are attached to the legs and protective spars of the prototype. We also fabricated a two-winged prototype (left) for comparison. The design of this robot was adapted from that of the RoboBee\cite{ma2012design} originally created at the \textit{Harvard Microrobotics Laboratory}. A U.S. one-cent coin indicates the scale. \label{FigRAL01}}
\end{center}
\vspace{-4ex}
\end{figure}

To achieve underactuated controllability, Harvard researchers developed the RoboBee, which is driven by two independent bimorph actuators \cite{ma2013controlled}. Dynamic analyses indicate that two-winged robots of this type should be able to perform basic flight maneuvers such as perching, landing, path following and obstacle avoidance by controlling the six spatial \textit{degrees of freedom} (DOF)\cite{doman2010wingbeat}. However, during real-time control experiments, it has been observed that the yaw torque produced via \textit{split-cycle} flapping \cite{doman2010wingbeat} is insufficient to overcome the restoring and damping forces opposing the yaw rotational motion of the robot~\cite{fuller2019four,fuller2015rotating,gravish2016anomalous}. A different approach to achieve controllability at the insect scale is the three-actuator flapping-wing design in~\cite{finio2012open}, which is composed of one central bimorph actuator employed for power and control, and two smaller lateral bimorph actuators used exclusively for control. Even though this robot can roll and pitch, the inability to steer itself has prevented it from performing agile flying maneuvers. 

Recently, following an approach that deviates from the bioinspiration paradigm, a 143-mg four-winged design was introduced in~\cite{fuller2019four}. This robot (dubbed Four-wings) is composed of four bimorph actuators configured horizontally to form a 90-degree cross, thus resembling the shape of a quadrotor. Due to its configuration, Four-wings exhibits a significantly-improved payload capacity compared to those of previous designs and can effectively steer itself, which suggests that it might be able to perform nontrivial controlled flying maneuvers. Note that these new capabilities directly follow from the fact that by increasing the number of actuators, the control authority is also increased (as the degree of underactuation decreases). This notion is clearly supported by research on larger-scale four-winged flying robots; for example, the DelFly Nimble \cite{karasek2018tailless} (with a weight of 29\,g and a wingspan of 330\,mm), which is equipped with two actuators for flapping, one actuator for dihedral-angle control and one actuator for wing-root control, is able to perform a large number of insect-inspired aerobatic maneuvers such as 360$^{\circ}$-flips and fast banked turns.

\begin{figure*}[t!]
\vspace{-2ex}
\begin{center}
\includegraphics[width=6.5in]{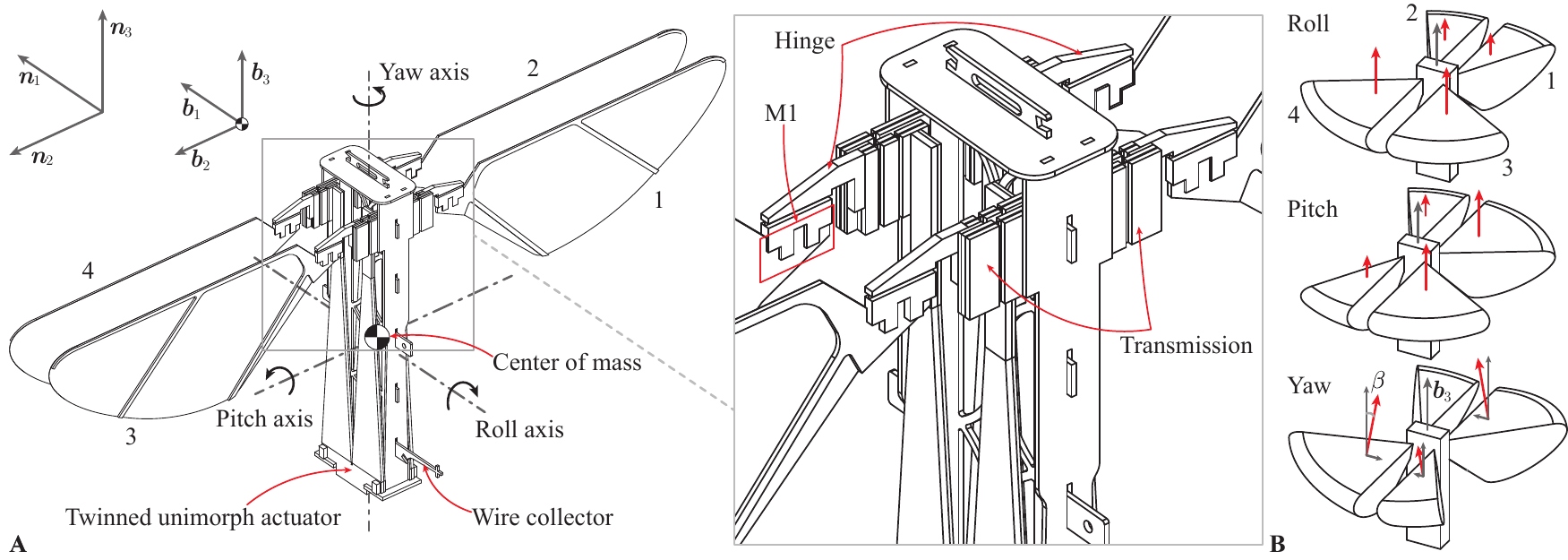}
\caption{Schematic diagrams of the four-winged robotic design. \textbf{A}.~This figure shows the inertial frame $\bs{n}_1$-$\bs{n}_2$-$\bs{n}_3$, the body frame $\bs{b}_1$-$\bs{b}_2$-$\bs{b}_3$ (shifted for clarity), the three body rotation axes as well as the numeric labels, $\left\{ 1,2,3,4\right\}$, that are utilized to indicate each wing and the respective unimorph actuator. A detailed view of the wings, hinges, transmissions is shown in the middle. \textbf{B}.~Strategies employed to generate the control torques about the three orthogonal axes of rotation. The roll torque is generated by varying the flapping-amplitude difference between the left wings $\left\{ 3,4 \right\}$ and the right wings $\left\{ 1,2 \right\}$. Similarly, by varying the flapping-amplitude difference between the front wings $\left\{ 2, 4 \right\}$ and the back wings $\left\{ 1, 3 \right\}$ the pitch torque is generated. As discussed in Section\,\ref{SECTION02B}, the steering motion about the yaw axis $\bs{b}_3$ can be produced employing three different methods \cite{roshanbin2019yaw,karasek2018tailless}. In this case, we only demonstrate the ISP method, which consists in pre-setting the stroke plane to have an inclination $\beta$ with respect to the steering plane $\bs{b}_1$-$\bs{b}_2$ and, then, by adjusting the flapping amplitudes of the diagonal pairs of wings, i.e., wings $\left\{ 1,4 \right\}$ and wings $\left\{ 2,3 \right\}$, the aerodynamic-force components projected on the $\bs{b}_1$-$\bs{b}_2$ plane produce the yaw torque while the cycle-averaged roll and pitch torques remain approximately zero. The flapping amplitudes are not shown to scale; the red arrows indicate the directions and magnitudes of the aerodynamic forces. \label{FigRAL02}}
\end{center}
\vspace{-2ex}
\end{figure*}

Here, motivated by the potential agility and controllability of flying robots with augmented actuation capabilities, we introduce Bee\textsuperscript{+}, a 95-mg insect-scale robotic design with four independently-driven wings powered by two pairs of twinned unimorph actuators (Fig.\,\ref{FigRAL01}). In this approach, rather than using four bimorph actuators as in~\cite{fuller2019four}, we employ two pairs of twinned unimorph actuators, as shown in~Fig.\,\ref{FigRAL02}, which are fabricated monolithically as shown in~Fig\,\ref{FIG03}. In the final assembly of the robot, each pair of twinned actuators is installed on each side of the airframe to independently drive the four wings of the system through four individual micro-transmissions as depicted in~Fig.\,\ref{FigRAL02}-A.
\begin{table*}[t!]
\caption{Comparison of the parameters of Bee\textsuperscript{+}, the RoboBee and Four-wings. \label{TABLE01}}
\vspace{-2ex}
\begin{center}
\begin{tabular}{c|c|c|c|c|c|c}
\hline
Robot & Total mass (mg) & Mass of the actuators (mg) & Wingspan (mm) & Flapping frequency (Hz) & Lift force (mN) & Wing area (\SI{}{\mm^2})\\
\hline
\hline
Bee\textsuperscript{+} & 95 & 56 & 33 & 100 & 1.4 & 200\\
\hline
RoboBee & 75 & 50 & 35 & 120 & 1.3 & 104\\
\hline
Four-wings & 143 & 100 & 56 & 160 & 4 & 218\\
\hline
\end{tabular}
\end{center}
\vspace{1ex}
\end{table*}

The main characteristics and parameters of Bee\textsuperscript{+}, compared to those of the RoboBee and Four-wings are summarized in Table\,\ref{TABLE01}. These data allows us to state the main characteristics of Bee\textsuperscript{+} in comparison to the best state-of-the-art insect-scale robots thus far presented in the technical literature: 

\hspace{-2ex}(i)\,The pair of twinned unimorph actuators weighs only 28\,mg, i.e., only 3\,mg more than a single bimorph actuator (25\,mg) used in the fabrication of the RoboBee and Four-wings. Consequently, the total weight of Bee\textsuperscript{+} is not significantly higher than that of the two-winged RoboBee and is lighter than that of the Four-wings.

\hspace{-2ex}(ii)\,Due to its compact configuration and the short wingspan (33\,mm), the volume of the fictitious parallelepipedal envelop enclosing Bee\textsuperscript{+} is almost identical to that of the RoboBee and significantly smaller than that of Four-wings, which has its four actuators oriented horizontally. 

\hspace{-2ex}(iii)\,The total wing area of Bee\textsuperscript{+} is twice as large as that of the RoboBee while its weight is only \SI{27}{\percent} higher, which significantly reduces the total \textit{wing-loading} on the robot. Lower wing-loading not only reduces the forces and moments acting on the robot's actuators, which increases the life-expectancy of the mechanical components, but it is also advantageous from the aerodynamic design viewpoint (see Section\,\ref{SECTION02B}).

\hspace{-2ex}(iv)\,The novel design of the proposed twinned unimorph actuators, fabricated using the methods described in\,\cite{jafferis2015design}, significantly reduces the complexity of the fabrication process and the statistical frequency of assembling errors compared to that of the two-winged robots. Also, compared to Four-wings, the circuitry of the four actuators driving Bee\textsuperscript{+} is simpler as it requires only five connection wires instead of six.  

In the rest of the paper, we first describe the design of the proposed four-winged robot and present a basic aerodynamic analysis of the system. Then, we discuss the fabrication process of the unimorph actuators employed to drive the robot. Next, we present the process of controller synthesis and a set of controlled flight experiments. Finally, we draw some conclusions extracted from the research results.

\vspace{-0.5ex}
\section{Design and Analysis}
\vspace{-0.5ex}
\label{SECTION02}
\vspace{-0.5ex}
\subsection{Robotic Design}
\vspace{-0.5ex}
\label{SECTION02A}
Although four-winged microrobotic flyers provide more options for the design and implementation of high-performance flight controllers, when compared to prototypes with two wings, their development brings numerous challenges. From the fabrication perspective, the integration of multiple actuators into a cm-scale airframe is difficult because the functionality and performance of microrobots greatly depend on the uniformity of the moving parts, symmetry of the structural components and precision of the final assembly. In addition, actuators are major contributors to the total weight of insect-sized robots (for example, approximately \SI{66}{\percent} of the weight of the robot in~\cite{Wood2008}); therefore, the addition of actuators to a robotic design requires the generation of significant more lift. We overcame these challenges by introducing an optimized method for the integration of four actuators into the robot.

As seen in~Fig.\,\ref{FigRAL01}, the most distinctive characteristic of Bee\textsuperscript{+} is its four-winged design compactly packaged inside a volume similar to that of the two-winged RoboBee. The robot has a symmetric configuration with respect to the $\bs{b}_1$--$\bs{b}_3$ plane that separates the right and left sides of the body frame of reference (as defined in Fig.\,\ref{FigRAL02}); in this case, by convention, wings\,1 and 2 are located in the right half-space and wings\,3 and 4 are located in the left half-plane. Each wing flaps only within its corresponding quadrant defined by the body $\bs{b}_1$-$\bs{b}_2$ plane, so less amplitude of deflection is required from each actuator compared to those in the two-winged robot case. The key element that makes this design feasible is the pair of twinned unimorph actuators with a common base that are shown in Fig.\,\ref{FigRAL02}-A. Note that each pair of twinned unimorph actuators can be thought of as an unfolded bimorph actuator, which explains why the weight difference between these two types of actuation microdevices is of 3\,mg only (the analogous bimorph actuator is 3\,mg lighter). In total, Bee\textsuperscript{+} is 20\,mg heavier than the RoboBee due to other additional structural weight; this is not an issue, however, as Bee\textsuperscript{+} is able to generate sufficient thrust and aerodynamic moments for flying and control.

Because the two pairs of twinned unimorph actuators are fabricated from the same composite stack and employing exactly the same process, their mechanical properties, functionalities and achieved performances are very similar. This fabrication methodology is simpler and more precise than pairing actuators as done in the case in~\cite{ma2012design}. Also, the use of twinned components eliminates the possibility of misalignment due to assembly errors on each side of the robot's body as each pair of unimorph actuators is a monolithic piece; thus, we only need to enforce the symmetry of the left side with respect to the right side. To power Bee\textsuperscript{+}, a minimum of five wires is required: two for the driving signals of each pair of twinned actuators and one for the common ground.
\vspace{-0.5ex}
\subsection{Aerodynamic Design and Analysis}
\vspace{-0.5ex}
\label{SECTION02B}
Bee\textsuperscript{+} has superior controllability capabilities compared to those exhibit by two-winged robots. However, there are two adverse factors that must be considered: the increased total weight of the robot; and the fact that each wing is constrained to flap with amplitudes equal or smaller than \ang{90} due to the geometry of the design, which limits the maximum thrust that each wing can generate. Namely, it is not a trivial task to guarantee the generation of sufficient thrust and moments to enable the robot to take off, stabilize itself and maneuver. For a wing flapping according to a sinusoidal pattern, the aerodynamic force, which depends on the velocity of the local flow and the corresponding angle of attack~\cite{chang2018time}, is the main contribution to the cycle-averaged lift
\begin{align}
\bar{f}_\text{L} &= C_\text{L}(\bar{\alpha}) \frac{1}{2}\rho (2\phi_0 \nu r_\text{ref})^2 S = C_1(\bar{\alpha}) \nu^2\phi_0^2 S
\label{EQ01}
\end{align} 
where $C_\text{L}$ is the cycle-averaged lift coefficient as a function of the aerodynamic mean angle of attack $\bar{\alpha}$; $\rho$ is the density of the air; $\phi_0$ is the end-to-end amplitude of the flapping angle; $\nu$ is the flapping frequency; $r_\text{ref}$ is the characteristic distance used to estimate the local velocity of the flow interacting with the wing; $S$ is the area of the wing; and $C_1$ is a lumped coefficient that simplifies the expression. 
 
The form of (\ref{EQ01}) indicates that, for control purposes, $\bar{f}_\text{L}$ can be modulated by either varying the frequency $\nu$ or the amplitude $\phi_0$. Note, however, that this formula provides a quick estimation only and, therefore, we employ the numerical fluid-structure interaction method in \cite{chang2016dynamics} and the instantaneous aerodynamic models in \cite{chang2018time} to compute the forces produced by the four wings of the robot and the corresponding total lift in~Table\,\ref{TABLE01}. For the purpose of design, we select $\nu = \SI{100}{Hz}$, $\phi_0 = \ang{65}$, the limit for the wing pitching angles to be \ang{70} and the hinge stiffness to be \SI{1.4}{\micro Nm}. For these parameters and the wing geometry shown in Fig.\,\ref{FigRAL02}-A, the computed  cycle-averaged total lift produced by the four wings is approximately \SI{1.4}{mN} and the corresponding lift-to-weight ratio is approximately 1.4; hence, based on this estimations, Bee\textsuperscript{+} is capable of generating sufficient lift for taking off, stabilizing itself and maneuvering.
  
Experiments have shown that two-winged robots are not well suited to passively resist rotational disturbances and actively steer their bodies about the yaw axis $\bs{b}_3$~\cite{fuller2019four}; phenomena that here we analyze using the cycle-averaged damping force, $\bar{f}_\text{D}$. This approach is reasonable because for insect-scale microrobots, the flapping frequency $\nu$ is significantly higher than the frequencies of the body oscillations \cite{chang2018time,fuller2019four}. Thus, for the upstroke and downstroke of a sinusoidal flapping pattern with a symmetrical profile, the cycle-averaged damping force \cite{cheng2011translational,chang2018time} is estimated as 
\begin{align}
\bar{f}_\text{D} = C_2(\bar{\alpha})  \phi_0 \nu \omega_\text{b} S + C_3(\bar{\alpha})  \dot{\omega}_\text{b} S
\label{EQ02}
\end{align}
where $C_2$ and $C_3$ are coefficients derived from the models in \cite{chang2018time}, and $\omega_\text{b}$ is the angular velocity associated with the yaw rotation. In (\ref{EQ01}), $\bar{f}_\text{L}$ is proportional to $\phi_0^2 \nu^2 S$; therefore, given similar robot weights and a constant $\omega_\text{b}$, from (\ref{EQ02}) it follows that Bee\textsuperscript{+} can generate at least $\sqrt{2}$~times the damping force produced by a two-winged robot of a similar size. This fact explains the enhanced yaw-stability properties of Bee\textsuperscript{+} when compared with the two-winged counterparts.    
 
In this discussion, we select the yaw \textit{steering plane} to be the $\bs{b}_1$-$\bs{b}_2$ body plane in Fig.\,\ref{FigRAL02}. Thus, to enable yaw steering capabilities, each wing must be able to actively generate a non-zero net force $f_\text{S}$ in the $\bs{b}_1$-$\bs{b}_2$ plane during one flapping cycle. From the conceptual design perspective, there are three feasible strategies available to generate a non-zero $f_\text{S}$. The first is \textit{split-cycle}. From simple analyses\cite{doman2010wingbeat} and experimental data obtained using the Four-wings prototype\cite{fuller2019four}, it follows that this strategy requires a high actuation bandwidth for both frequency modulation and yaw-torque amplification, which is costly and difficult to achieve from the design and fabrication perspective. The second is \textit{asymmetric angle of attack}. This is the method employed by the DelFly Nimble in \cite{karasek2018tailless}, which uses an actuator to actively control the wing root; in this way, the angles of attack of the wing during the up and down strokes can be set to different values. The third is \textit{inclined stroke-plane} (ISP), which is employed in this case to control the yaw DOF of Bee\textsuperscript{+}. This method consists in pre-setting the stroke plane to have an inclination ($\beta$ in Fig.\,\ref{FigRAL02}-B) with respect to the steering plane. Specifically, the stroke planes of the front pair of wings $\left\{ 2,4 \right\}$ are tilted backward while that of the back pair if wings $\left\{ 1,3 \right\}$ are tilted forward. In this way, the aerodynamic force produced by a wing projects a non-zero component onto the steering plane. 

According to the ISP scheme, the diagonal pair $\left\{ 1,4 \right\}$ can produce yaw torques in the counter-clockwise direction while the other diagonal pair $\left\{ 2,3 \right\}$ can generate clockwise yaw torques. Thus, by adjusting the flapping amplitudes of the two diagonal pairs of wings, the robot can actively modulate the production of yaw torque. In specific, the active yaw torque generated by the projection of the cycle-averaged lift produced by a single wing onto the steering plane can be estimated as
\begin{align}
\bar{f}_\text{S}  &= \bar{f}_\text{L}\sin\beta = C_4(\bar{\alpha})  \nu^2\phi_0^2  \label{EQ03}\\
\bar{\tau}_\text{S} &= r_\text{S} \bar{f}_\text{S} = C_5(\bar{\alpha})  \nu^2\phi_0^2      \label{EQ04}
\end{align}
where $r_\text{S}$ is the distance from the pressure center of the wing to the $\bs{b}_3$ axis, and $C_4$ and $C_5$ are lumped coefficients derived from the models for instantaneous forces described in~\cite{chang2018time}. Note that unlike in the two-winged case, by diagonally pairing the four wings, the modulation of the yaw torque does not introduce significant undesirable roll torques as they stay approximately balanced. 
 
Since the torques about the three axes of the body frame can be controlled by varying the flapping amplitudes of the four wings (see Fig.\,\ref{FigRAL02}-B), the robot can be controlled during flight employing methods already developed for quadrotors~\cite{Ying2018IROS,chen2017lyapunov}. Unfortunately, since $\bar{\alpha}$ can vary along with $\phi_0$ in a highly nonlinear manner, the models specified by (\ref{EQ01})--(\ref{EQ04}) cannot be used directly; with proper identification, however, they can be approximated with constant-coefficient linear models, as done in Section\,\ref{SECTION04D} for flight controller synthesis. Finally in this section, it is important to state that compared to the case of two-winged robots, the wing-loading on Bee\textsuperscript{+} is reduced by \SI{34}{\percent}. Lower wing-loading not only reduces the demands on the actuators but also induces smaller deformations of the wings and enables the generation of larger maximum flapping amplitudes, which is desirable for both power and control purposes. For example, in the two-winged case, the typical operating flapping amplitudes oscillate around \ang{110} while the designed value of $\phi_0$ for Bee\textsuperscript{+} can be chosen to be significantly larger than \ang{55}. In static experiments (see supplementary movie S1.mp4), the maximum observed amplitudes achieve values of approximately \ang{75}.
\begin{figure*}[t]
\begin{center}
\includegraphics[width=6.65in]{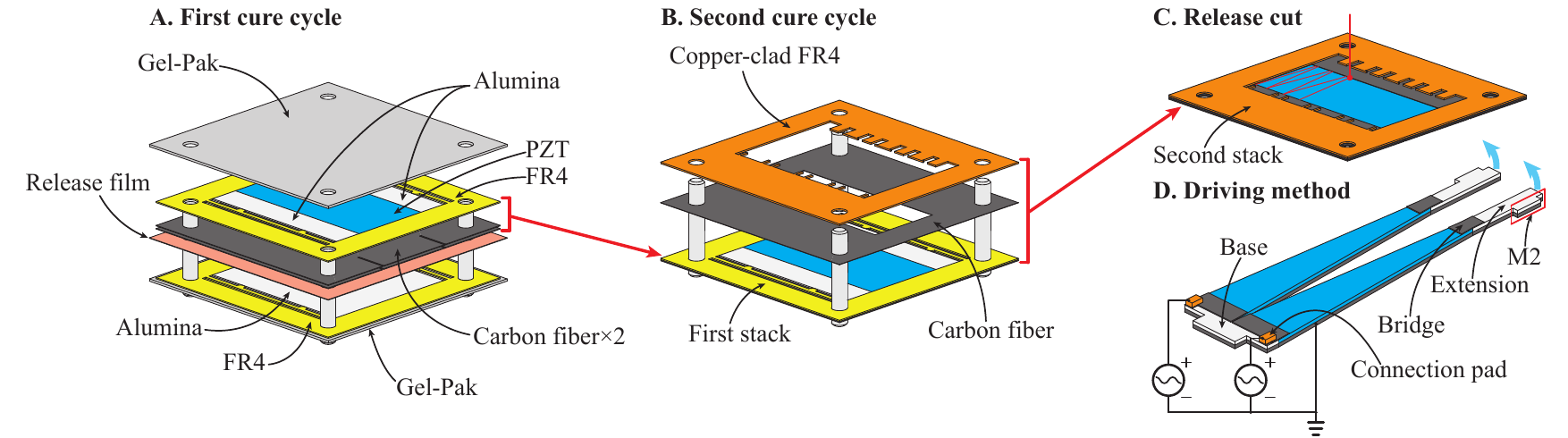}
\end{center}
\caption{Fabrication process of a batch of twinned piezoelectric unimorph actuators. \textbf{A.}~During the \textit{first cure cycle}, rectangular laminates of PZT (\SI{127}{\um}) and alumina (\SI{127}{\um}) are bonded to two layers of high-modulus carbon fiber composite (\SI{63}{\um} each) by applying heat (\SI{180}{\degreeCelsius}) and pressure (\SI{15}{psi}) to the stack, which is pin-aligned. An additional bottom layer of alumina serves as a substrate that maintains the stack flat; between this piece of alumina and the double layer of carbon fiber we place a sheet of release film to prevent undesired bonding. \textbf{B.}~During the \textit{second cure cycle}, the cured unimorph stack produced in the first step is bonded to an additional layer of carbon fiber composite (\SI{27}{\um}) and a layer of copper-clad FR4 (\SI{137}{\um}) using the same temperature and pressure than in the first cycle. \textbf{C.}~During the final \textit{release cut}, four pairs of twinned actuators are released from one unimorph stack. \textbf{D.}~The \textit{driving method} employs two independent voltages per each pair of unimorph actuators. In this case, we use positive sinusoidal signals with maximum magnitudes of \SI{260}{V}; accordingly, the two unimorphs bend upwards (as indicated by the arrows), generating maximum tip displacements of \SI{300}{\um}. The total length of a twinned pair is \SI{13}{mm}. \label{FIG03}}
\vspace{-2ex}
\end{figure*}

\vspace{-0.5ex}
\section{Fabrication of Unimorph Actuators}
\vspace{-0.5ex}
\label{SECTION03}
The proposed twinned unimorph actuators are the most important components of the mechanisms that flap the four wings of the robot (Fig.~\ref{FigRAL02}-A). The fabrication of actuators of this type is not feasible employing the methods invented to create the bimorph actuators used to drive the RoboBee in ~\cite{ma2012design}. Here, we introduce a new technique that is based on a modification of the \textit{pre-stack} technology described in~\cite{jafferis2015design}. This approach not only significantly improves the quality and consistency of the fabrication process but also enables the physical realization of almost any planar design. The specific process employed in the construction of the unimorph actuators of the Bee\textsuperscript{+} prototype in~Fig.\,\ref{FigRAL01} is depicted in Fig.\,\ref{FIG03}, which consists of two cure cycles (Figs.\,\ref{FIG03}-A~and~\ref{FIG03}-B) and one laser release cut (Fig.~\ref{FIG03}-C). We use piezoelectric ceramics PZT-5H (T105-H4NO-2929, Piezo.com) as the \textit{active layer} to create the unimorph structure because of its high-modulus and piezoelectric coefficient, and two layers of high-modulus carbon fiber composite as the \textit{passive surface constraint} in order to obtain an equivalent stiffness similar to that of a bimorph actuator of the same size. The tip extension and base are made of alumina ceramics.

Before the first cure cycle starts, we cut all the pieces of laminated materials required to assemble the first stack in Fig.\,\ref{FIG03} using a precision \textit{diode-pumped solid-state} (DPSS) laser (Photonics Industries DC150-355). The pieces of PZT and alumina are first cut into rectangles, and then cleaned with isopropyl alcohol using a sonicator in order to improve the adhesion between layers during curing. Sheets of FR4 are machined as jigs to hold the pieces of PZT and alumina in position, thus forming two layers of the pin-aligned stack as shown in Fig.~\ref{FIG03}-A. Here, the lower FR4 jig is placed on a layer of Gel-Pak (lightly tacky film) that is used to stick the piece of alumina in place. Similarly, the upper FR4 jig, which holds the pieces of PZT in place, is covered with a layer of Gel-Pak. As shown in Fig.\,\ref{FIG03}-A, from the bottom up, the stack also contains a layer of release film and two layers of carbon fiber. Through the application of heat (\SI{180}{\degreeCelsius}) and pressure (\SI{15}{psi}) for two hours, the epoxy resin impregnated in the two layers of carbon fiber cures and bonds these pieces together, and them with the layer composed of PZT, alumina and FR4, thus forming the first cured stack employed to assemble the second stack that is shown in Fig.\,\ref{FIG03}-B. Note that the lower layer of the first stack serves only as a rigid substrate that maintains the configuration flat; therefore, it does not bond to the carbon fiber pieces as a layer of release film isolates them from each other. During the second cure cycle (Fig.\,\ref{FIG03}-B), we apply the same temperature and pressure patterns to bond the first cured stack with an additional layer of carbon fiber and a copper-clad FR4 sheet. This last layer of carbon fiber \textit{structurally bridges} the interfaces between the pieces of PZT and alumina, thus increasing the rigidity of the actuator. The layer of copper-clad FR4 is necessary to make the electrical connection pads.

Finally in the fabrication process, the unimorph actuators are released from the second cured stack through a final laser cut. As the final stack has only one layer of PZT at the top and one layer of carbon fiber at the bottom, during the final release procedure, we simply cut all the layers at once from the top as shown in Fig.\,\ref{FIG03}-C. This simpler laser cut reduces the releasing time by half compared to that required to release bimorph actuators, as the final fabrication step of bimorphs consists of one cut from the top and another cut from the bottom of the corresponding stack. The one-single-cut-based final release procedure of the unimorph fabrication method significantly improves the yield of actuators per stack. This observation is explained, to some extent, by the fact that the final release of bimorphs requires the flipping of the cured stack and its realignment in between the top and bottom laser cuts, which increases the likelihood of introducing manufacturing errors such as the induction of cracks in the PZT ceramics.

In this case, we apply the fabrication method in such a way that from one stack, the final release cut yields a total of four pairs of twinned actuators. One twinned pair is depicted in Fig.\,\ref{FIG03}-D. As seen in this illustration, two identical unimorphs located side by side share the same structural base and electrical ground. At the base, there is a clearance of \SI{50}{\um} that separates the bottom edges of the PZT layers of both twined 
unimorphs. In contrast, their passive carbon fiber layers are connected at the bottom, thus creating a common electrical ground. Consistently, to drive the actuator pair, we can employ the simple circuitry shown in Fig.~\ref{FIG03}-D. In this electrical configuration, two independent voltage signals drive the two unimorphs, whose bending directions when excited are indicated with \textit{blue} arrows. To connect the electrical copper pads with the PZT layers, we employ conductive epoxy.
 
\vspace{-0.5ex} 
\section{Flight Controller Design}
\vspace{-0.5ex}
\label{SECTION04}
\vspace{-0.5ex}
\subsection{System Dynamics}
\vspace{-0.5ex}
\label{SECTION04A}
To describe the dynamics of the robot, we define the body-fixed frame $\bs{b}_1$-$\bs{b}_2$-$\bs{b}_3$ \big($\bs{\mathcal{B}}$\big) and the inertial frame $\bs{n}_1$-$\bs{n}_2$-$\bs{n}_3$ \big($\bs{\mathcal{N}}$\big), as shown in Fig.~\ref{FigRAL02}. Because the direction of the thrust force is assumed to be aligned with the $\bs{b}_3$ axis and the number of actuators is less than the total number of the DOF of the system, Bee\textsuperscript{+} is essentially a thrust-propelled underactuated system. Thus, it can be thought of as a rigid body with its dynamics given by
\begin{align}
m \ddot{\bs{r}}
&= -mg \bs{n}_3 + f \bs{b}_3 \label{EQ05} \\
\bs{J} \dot{\bs{\omega}} &= -\bs{\omega} \times \bs{J} \bs {\omega} + \bs{\tau} \label{EQ06} \\
\dot{\bs{\qbar}}& = \frac{1}{2}\bs{\qbar}\ast\bs{\pbar} \label{EQ07}  
\end{align} 
where $m$ is the total mass of the robot; $\bs{r} = \left[r_{1}\;r_{2}\;r_{3}\right]^T$ indicates the location of the robot's center of mass measured from the origin of $\bs{\mathcal{N}}$; $f$ is the magnitude of the total thrust force generated by the four flapping wings; $\bs{J}$ denotes the robot's moment of the inertia, written with respect to $\bs{\mathcal{B}}$; $\bs{\omega}$ is the flyer's angular velocity with respect to $\bs{\mathcal{N}}$, expressed in $\bs{\mathcal{B}}$; $\bs{\tau}=\left[\tau_1\;\tau_2\;\tau_3\right]^T$ is the torque generated by the flapping wings; the quaternion $\bs{\qbar}$ describes the attitude of the robot relative to $\bs{\mathcal{N}}$; $\bs{\pbar} = \left[0 \;\; \bs{\omega}^T\right]^T$; and the symbol $\ast$ denotes the standard quaternion multiplication.
 
Note that the model specified by (\ref{EQ05})--(\ref{EQ07}) assumes that the direction of the thrust force is aligned with $\bs{b}_3$; that the projection of the total aerodynamic force generated by the four flapping wings onto the steering plane is zero during one flapping cycle, which implies that $f\bs{b}_3$ is the only external actuation force acting on the system; that the aerodynamic disturbances affecting the flyer are negligible; and that the gyroscopic effect resulting from the interaction of the flapping wings with the rotating body is also negligible.
  
\vspace{-0.5ex}  
\subsection{Attitude Control}
\vspace{-0.5ex}
\label{SECTION04B}
Here, we describe the desired attitude dynamics of the robot with the quaternion equation 
\begin{align}
\bs{\dot{\qbar}}_{\ts{d}} = \frac{1}{2}\bs{\qbar}_{\ts{d}}\ast\bs{\pbar}_{\ts{d}} \label{EQ08}
\end{align}
where $\bs{\qbar}_{\ts{d}}$ is the quaternion employed to represent the desired attitude of the flyer during flight; and $\bs{\pbar}_{\ts{d}} = \left[0\;\;\bs{\hat{\omega}}_{\ts{d}}^T\right]^T$, in which $\bs{\hat{\omega}}_{\ts{d}}$ denotes the desired angular velocity expressed in the desired frame of reference, whose orientation coincides exactly with that of $\bs{\qbar}_{\ts{d}}$ (see \cite{chen2017lyapunov} for further details). Consistently, it follows that the attitude error between $\bs{\qbar}_{\ts{d}}$ and $\bs{\qbar}$, described by the quaternion $\bs{\qbar}_{\ts{e}} = \left[m_\ts{e} \;\; \bs{n}_{\ts{e}}^T\right]^T$, is given by 
\begin{align}    
\bs{\qbar}_{\ts{e}} = \bs{\qbar}_{\ts{d}}^{-1}\ast\bs{\qbar}. \label{EQ09}
\end{align}

In this case, we regulate the attitude of the flyer employing the control law  
\begin{align}
\bs{\tau} = -\bs{K}_1\ts{sgn}(m_{\ts{e}})\bs{n}_{\ts{e}} - \bs{K}_2(\bs{\omega} - \bs{\omega}_{\ts{d}}) \label{EQ10}
\end{align}
in which $\bs{K}_1$ and $\bs{K}_2$ are positive definite diagonal gain matrices; $\ts{sgn}(\cdot)$ denotes the sign function; and $\bs{\omega}_{\ts{d}}$ is the desired angular velocity with exactly the same components as those of $\bs{\hat{\omega}}_{\ts{d}}$ but expressed in the body frame instead of the desired frame with the same orientation as that of $\bs{\qbar}_{\ts{d}}$. We represent the axis of the rotation from $\bs{\qbar}$ to $\bs{\qbar}_{\ts{d}}$ with the unit vector $\bs{a}_{\ts{e}}$ and we use $\Theta_{\ts{e}}$ to denote the associated rotation angle, with $0 \leqslant \Theta_{\ts{e}} <\pi$. Thus, note that the term $-\ts{sgn}(m_{\ts{e}}) \bs{n}_{\ts{e}}$ is geometrically identical to $\sin \hspace{-0.4ex} {\left( \Theta_{\ts{e}} / 2 \right)} \hspace{0.1ex} \bs{a}_{\ts{e}}$ and that we use the multiplier $\ts{sgn}(m_{\ts{e}})$ to remove the ambiguity associated with the quaternion notation according to which both $\bs{\qbar}_{\ts{e}}$ and $-\bs{\qbar}_{\ts{e}}$ represent the same rotation.

\vspace{-0.5ex}
\subsection{Position Control}
\vspace{-0.5ex}
\label{SECTION04C}
To control the position of the robot in space, we employ as control signals the magnitude and direction of the total thrust generated by the flapping wings. According to this approach, the magnitude of the total thrust force, $f$, is modulated by jointly varying the flapping speed of the four wings and the corresponding force direction is modulated by varying the attitude of the robot. As described in Section\,\ref{SECTION04B}, the flyer's attitude is controlled with the feedback law specified by (\ref{EQ10}), which is physically realized by flapping the wings asymmetrically according to the patterns depicted in Fig.\,\ref{FigRAL02}-B. In this case, the \textit{desired} instantaneous total thrust force required to track a desired position of the flyer's center of mass, $\bs{r}_\ts{d}$, is generated according to the \textit{proportional-integral-derivative} (PID) structure 
\begin{align}
\begin{split}
\bs{f}_{\ts{d}} =& -\bs{K}_{\ts{p}}(\bs{r} - \bs{r}_{\ts{d}}) - \bs{K}_{\ts{d}}(\bs{\dot{r}} - \bs{\dot{r}}_{\ts{d}}) \\
& -\bs{K}_{\ts{i}}\int(\bs{r} - \bs{r}_{\ts{d}}) dt  + mg\bs{n}_{3} + m\bs{\ddot{r}}_{\ts{d}} \label{EQ11}
\end{split}
\end{align}
where $\bs{K}_{\ts{p}}$, $\bs{K}_{\ts{d}}$ and $\bs{K}_{\ts{i}}$ are positive definite diagonal gain matrices. Note that the magnitude of $\bs{f}_{\ts{d}}$ to be tracked using direct feedback control is simply given by 
\begin{align}
f_{\ts{d}} =& \bs{f}_{\ts{d}}^T\bs{b}_3. \label{EQ12}
\end{align}

The set of all the flyer's attitudes compatible with the direction of $\bs{f}_{\ts{d}}$ can be readily computed as 
\begin{align}
\bs{i}_{3} = \frac{\bs{f}_{\ts{d}}}{\left|\bs{f}_{\ts{d}}\right|_2} \label{EQ13}
\end{align}
which is chosen to be the desired yaw axis of the robot during flight. We compute the other two axes defining the desired attitude of the robot in terms of the desired instantaneous yaw rotation angle, $\psi_{\ts{d}}$, and $\bs{i}_{3}$, according to
\begin{align}
\bs{i}_{1} = \frac{{\bs{i}_{\psi}}_{\ts{d}} \times \bs{i}_3} {\left| {\bs{i}_{\psi}}_{\ts{d}} \times \bs{i}_3 \right|_2},\;\;\;\;
\bs{i}_{2} =  \bs{i}_{3} \times \bs{i}_{1} \label{EQ14}
\end{align}
where ${\bs{i}_{\psi}}_{\ts{d}} = \left[-\sin{\psi_{\ts{d}}}\; \cos{\psi_{\ts{d}}}\; 0\right]^T$. In the implementation of the algorithms for signal processing and control, $\bs{i}_1$, $\bs{i}_2$, $\bs{i}_3$ and ${\bs{i}_{\psi}}_{\ts{d}}$ are expressed in the inertial frame. To implement the controller specified by (\ref{EQ10}), we compute the desired attitude quaternion $\bs{\qbar}_{\ts{d}}$ from the desired rotation matrix $\bs{S}_{\ts{d}} = \left[\bs{i}_1\; \bs{i}_2\; \bs{i}_3\right]$, employing standard quaternion algebra.

\vspace{-0.5ex}
\subsection{Actuator Command Generation}
\vspace{-0.5ex}
\label{SECTION04D}
As discussed in Section\,\ref{SECTION02B}, each actuator generates a sinusoidal output with a constant pre-specified frequency (100\,Hz in this specific case) and an adjustable amplitude used to generate the flapping patterns in Fig.\,\ref{FigRAL02}-B. By simplifying the models described by (\ref{EQ01})--(\ref{EQ04}), we estimate the magnitude of the thrust force produced by each flapping wing $j = \left\{1,2,3,4 \right\}$, according to $f_{\ts{j}} = k_f v_{\ts{j}}$, where $v_{\ts{j}}$ is the amplitude of the sinusoidal command signal generated by the \textit{j}th unimorph actuator and $k_f$ is a lumped thrust force coefficient. As illustrated in Fig.\,\ref{FigRAL02}-B, yaw torques in the steering plane can be generated by employing the ISP strategy discussed in Section\,\ref{SECTION02B}. Consistently, we estimate the component of the $i$th aerodynamic force projected on the steering plane as ${f_{\ts{s}}}_{\ts{j}} = k_s v_{\ts{j}}$, where $k_s$ is also a lumped force coefficient. 

Thereby, the mapping that relates the amplitudes of the actuators' outputs, as inputs, with the total thrust force and control torques, as outputs, is given by    
\begin{align}
\bs{u} = \bs{\Gamma} \bs{v}
\label{EQ15}
\end{align}
with 
\begin{align*}
\bs{u} &= 
\begin{bmatrix}
f & \tau_1 & \tau_2 & \tau_3 
\end{bmatrix}^T \\
\bs{\Gamma} &=
\begin{bmatrix}
     k_f          & k_f            &  k_f           &  k_f        \\
    -k_f d_1   & -k_f d_1   &  k_f d_1    &  k_f d_1  \\
     k_f d_2   & -k_f d_2   &  k_f d_2    & -k_f d_2  \\
     k_s d_3  & -k_s d_3  & - k_s d_3  & k_s d_3
\end{bmatrix}\\
\hspace{-8ex} \ts{and}~~~~~~~~
\bs{v} &= 
\begin{bmatrix}
v_1 & v_2 & v_3 & v_4 
\end{bmatrix}^T 
\end{align*}
where $d_{j}$, for $j = \left\{ 1,2,3\right\}$, is the equivalent lever-arm associated with the corresponding torque component $\tau_j$, employed to model, in an extremely simplified manner, the transmission mechanism that connects the actuators's output with the flapping angle of the $j$th wing. Thus, for a known set of control signals $\left\{ f, \tau_1, \tau_2, \tau_3 \right\}$, the corresponding set of instantaneous actuator commands is straightforwardly computed as $\bs{v} = \bs{\Gamma}^{-1} \bs{u}$. Note that, simply due to its four-winged design, Bee\textsuperscript{+} has better control capabilities than those of two-winged prototypes, because the thrust force and control torques are generated by four wings rather than two. 

\vspace{-0.5ex}
\section{Experimental Results}
\vspace{-0.5ex}
\label{SECTION05}
\subsection{Experimental Setup}
\vspace{-0.5ex}
\label{SECTION05A}
The main components of the experimental setup are a Bee\textsuperscript{+} prototype, four piezo-actuator drivers (PiezoMaster VP7206), a \textit{Vicon motion capture} (VMC) system and a ground target--host Mathworks Simulink Real-Time system that is used to process sensor measurements and generate the control signals. The control algorithms are run at a frequency of \SI{2}{\kilo\hertz} and the VMC system measures the robot's position and attitude at a rate of \SI{500}{\hertz}. The robot's angular velocity cannot be directly measured with the VMC system, thereby we estimate it according to
\begin{align}
\begin{bmatrix}
0 \\
\bs{\omega} 
\end{bmatrix}
= 2\bs{\qbar}^{-1}\ast\left[\frac{\lambda s}{s + \lambda}\right]\bs{\qbar} \label{EQ17}
\end{align}
where $s$ is the differential operator; the bracketed function on the right side represents a low-pass filter that operates on the signal $\bs{\qbar}$; and $\lambda$ is tuned filter parameter. To estimate the translational velocities, we employ a simple discrete-time differentiator in combination with a low-pass derivative filter similar to that in (\ref{EQ17}). Note that, in the case of Bee\textsuperscript{+}, the use of low-pass filters is necessary to clean the measured signals because the forces generated by the flapping wings induce high-frequency oscillations on the robot's body. Furthermore, the open-loop trimming flight tests required for controller tuning in the case of two-winged robots \cite{ma2013controlled,chirarattananon2014adaptive}, are not necessary in the case of Bee\textsuperscript{+} prototypes. This fact demonstrates that the proposed actuation and control methods do not require the fine tuning of the control signals and the zeroing of the biases affecting the actuation torques. This advantage has significantly improved the efficiency of flight experiments at the insect-scale.

\vspace{-0.5ex}
\subsection{Simultaneous Control of Altitude and Attitude}
\vspace{-0.5ex}
\label{SECTION05B}
The objective of the controller tested through this experiment is to enable the robot to fly at a desired altitude and with the direction of the thrust force remaining perpendicular to the $\bs{n}_1$-$\bs{n}_2$ plane. In this case, yaw feedback control is not employed in order to alleviate the actuation burden on the flapping wings and, also, to demonstrate that the proposed four-wing design significantly increases the aerodynamic damping along the yaw angular motion, which improves the open-loop stability of the yaw degree of freedom, as discussed in Section\,\ref{SECTION02B}. Theoretically, in the control of the thrust force direction, which coincides with the $\bs{b}_3$ axis, the Euler yaw angle can be ignored. Therefore, to simultaneously control the robot's altitude and attitude, we regulate the vertical coordinate of the body's center of mass to a desired constant, and we regulate the robot's roll and pitch angles to zero.

Accordingly, it follows that the attitude quaternion required to achieve the simultaneous experimental control of the robot's altitude and attitude is given by $\bs{\qbar}_{\ts{d}} =\left[\cos{(\psi/2)}\;\;0\;\;0\;\;\sin{(\psi/2)}\right]^T$, where $\psi$ is the actual Euler yaw angle according to the  Z-Y-X convention. In addition, the altitude controller can be simply derived from (\ref{EQ11}) and (\ref{EQ12}). In specific, under the assumption that $\bs{b}_3 \approx \bs{i}_3$, we immediately obtain that
\begin{align}
    f = -k_{\ts{p}}(r_{\ts{3}} - {r_{\ts{d}}}_{\ts{3}}) - k_{\ts{d}}\dot{r}_{\ts{3}} - k_{\ts{i}}\int(r_{\ts{3}} - {r_{\ts{d}}}_{\ts{3}})dt + mg
\end{align}
where $r_3$ is the measured altitude of the robot; ${r_{\ts{d}}}_{\ts{3}}$ is the desired altitude of the robot; and $k_{\ts{p}}$, $k_{\ts{d}}$ and $k_{\ts{i}}$ are controller gains found through classical control methods.
\captionsetup[subfigure]{labelformat=empty}
\begin{figure}[t!]
\begin{center}
\subfloat[\textbf{A.}~Reference ${r_\ts{d}}_3$ and measured altitude $r_3$.~~~~]{
\includegraphics[width = 7.4cm]{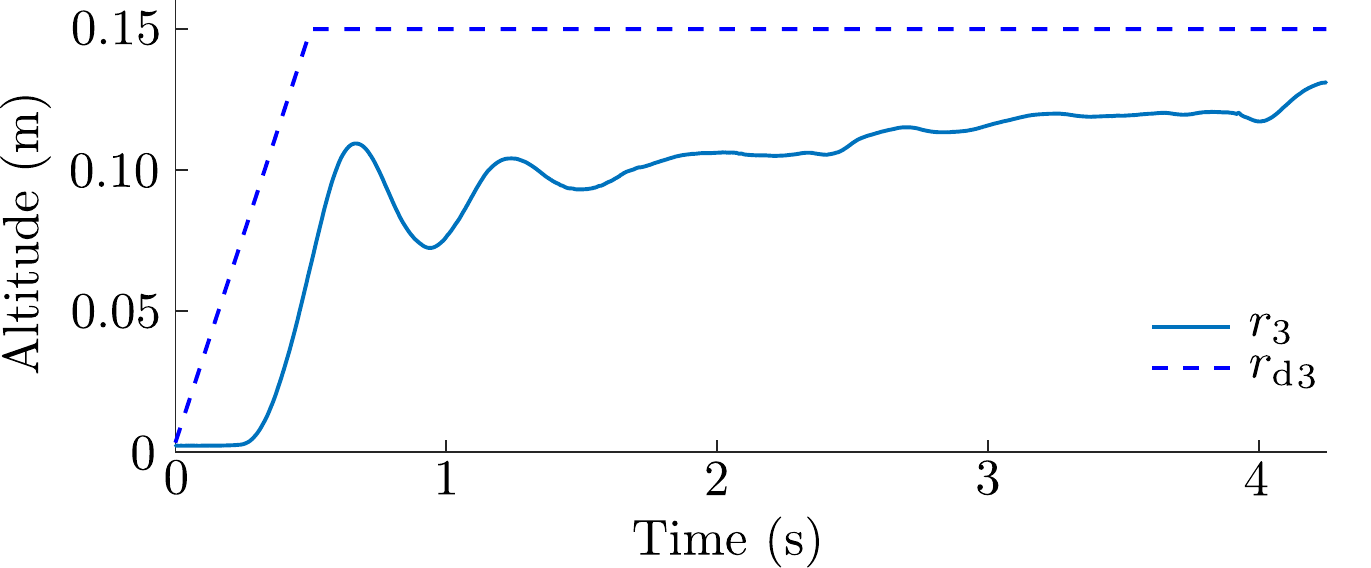}}
\hspace{10ex}
\subfloat[\textbf{B.}~Measured Euler roll and pitch angles.~~~~~~~~]{
\includegraphics[width = 7.4cm]{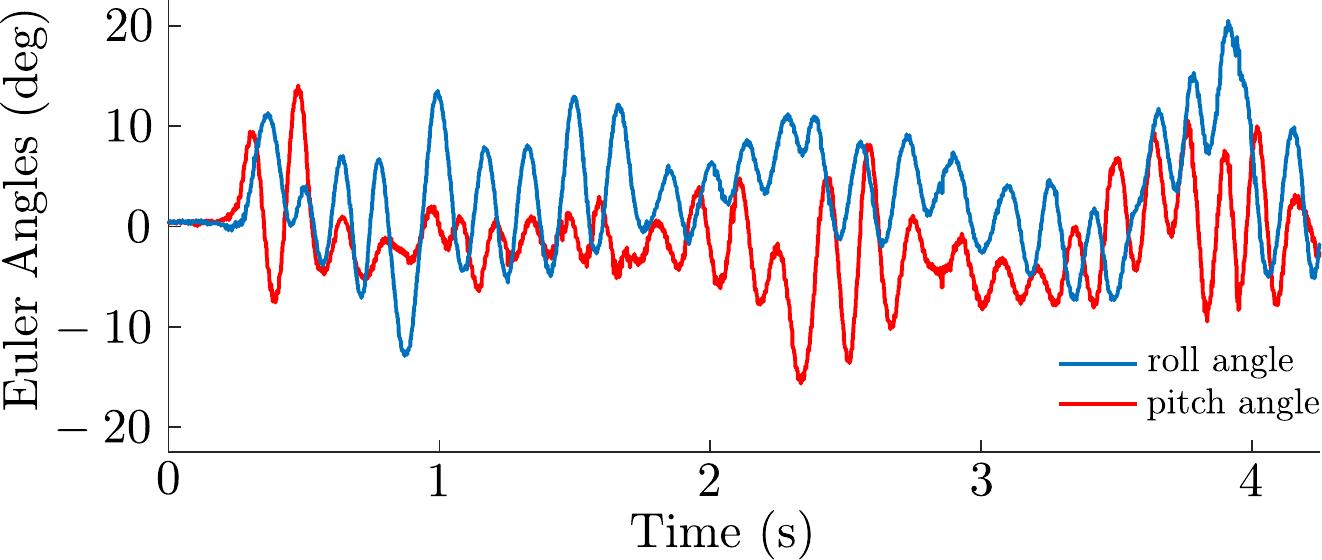}}
\caption{Simultaneous real-time control of altitude and attitude. \textbf{A}.~This plot shows that the measured altitude signal, $r_3$, tracks the main trend of the reference signal, ${r_\ts{d}}_3$; however, significant transient and steady-state errors can be observed. \textbf{B}.~This plot shows that the Euler roll and pitch angles oscillate approximately between \SI{-10}{\degree} and \SI{10}{\degree}, which is partially caused by the vibration of the robot's body that is induced by the flapping of the wings. The entire experiment lasts for approximately \SI{5}{\second}; then, the robot leaves the volume of operation and the power is automatically turned off. \label{FIG01}}
\end{center}	    
\vspace{-2ex}
\end{figure}
\begin{figure*}[t!]
\begin{center}
\includegraphics[width=7.1in]{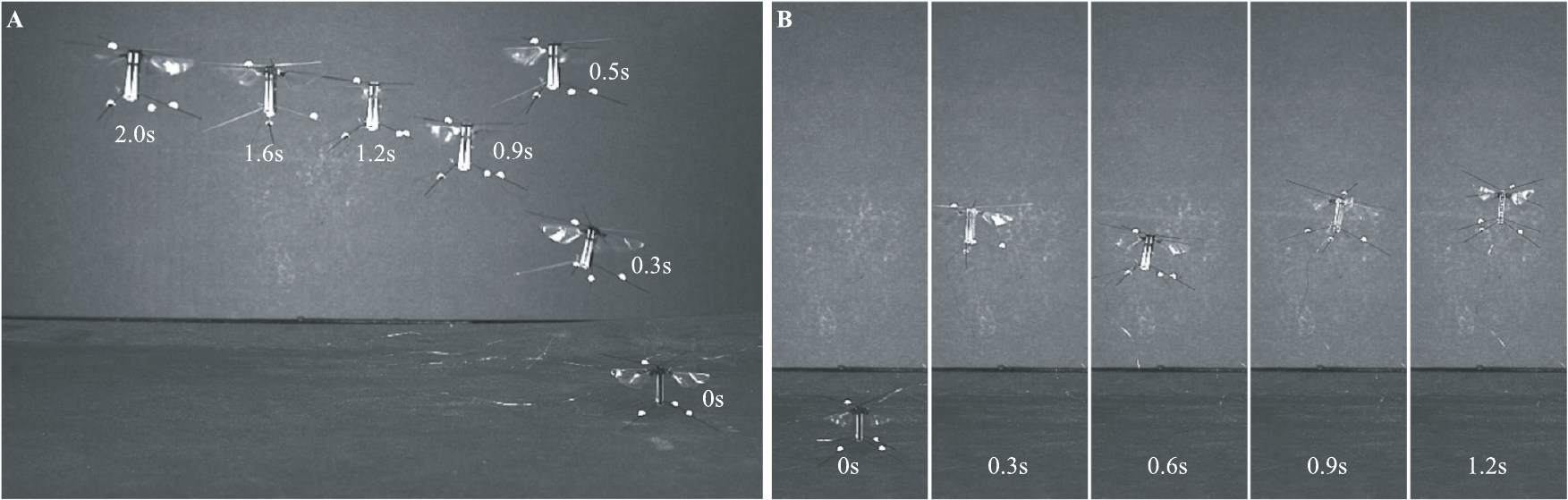}
\end{center}
\caption{\textbf{A}.~Photographic sequence of a flight experiment during which the altitude and altitude of the Bee\textsuperscript{+} prototype are simultaneously controlled. The corresponding altitude and Euler angles are shown in Fig.\,\ref{FIG01}. During the experiment, the direction of the thrust force is controlled to remain approximately perpendicular to the $\bs{n}_1$-$\bs{n}_2$ plane. The cable tethered to the robot provides the power and transmits the control signals. The robot drifts on the $\bs{n}_1$-$\bs{n}_2$ plane due to the lack of control actions along the horizontal inertial axes. After \SI{2}{\second}, the robot flies outside the focus area of the camera. \textbf{B}.~Photographic sequence of the position control experiment. The corresponding position and Euler angles of this experiment are shown in Fig.\,\ref{FigRAL06}. The complete set of experiments is shown in the supplementary movie S1.mp4. \label{FigRAL05}}
\vspace{-3ex}
\end{figure*}

A set of results obtained from an experiment in which the robot's altitude and attitude are simultaneously controlled is shown in Fig.\,\ref{FIG01}. Here, Fig.\,\ref{FIG01}-A compares the desired and measured altitudes, and Fig.\,\ref{FIG01}-B shows the measured roll and pitch angles of the robot for references equal to zero. From these data, it is clear that the algorithm specified by (\ref{EQ10}) is effective in controlling the robot's attitude as the experimental direction of the thrust force is approximately perpendicular to the $\bs{n}_1$-$\bs{n}_2$ plane. In this case, the angular oscillations stay mostly inside the range $\left[-10^\circ:10^\circ \right]$, which is acceptable in the sense that the magnitude of the lift force is not greatly degraded, as can be deduced from Fig.\,\ref{FIG01}-A. In specific, the robot rapidly reaches a value close to that of the desired altitude, even though the steady-state error does not seem to approach zero, probably due to an insufficient integral action. Furthermore, the time lapse of the experiment in Fig.\,\ref{FigRAL05}-A (also shown in the supporting movie S1.mp4) indicates that the four flapping wings generate a lift force sufficient for the robot to take off and maintain its body in the upright orientation for a significant period of time. The observed drifting phenomenon is expected due to the lack of control action in the $\bs{n}_1$-$\bs{n}_2$ plane. Overall, the data in Fig.\,\ref{FIG01} and photographic sequence in Fig.\,\ref{FigRAL05}-A provide compelling evidence demonstrating the effectiveness of the design, fabrication, actuation and control methods developed to create Bee\textsuperscript{+}. 
 
\vspace{-0.5ex} 
\subsection{Position Control Experiment}
\begin{figure}[t!]
\vspace{2ex}
\centering
\subfloat[~~~~~~\textbf{A.}~Reference and measured position of the center of mass.]{
\includegraphics[width = 7.4cm]{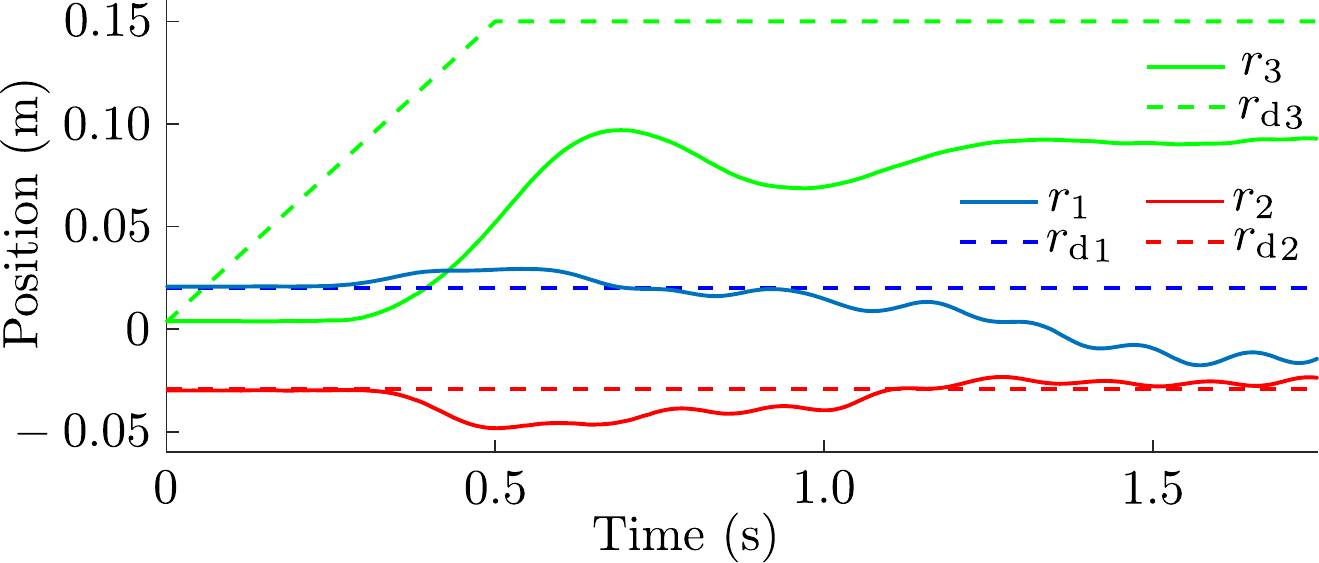}}
\hspace{10ex}
\subfloat[\textbf{B.}~References and measured roll and pitch angles.~~~~~]{
\includegraphics[width = 7.4cm]{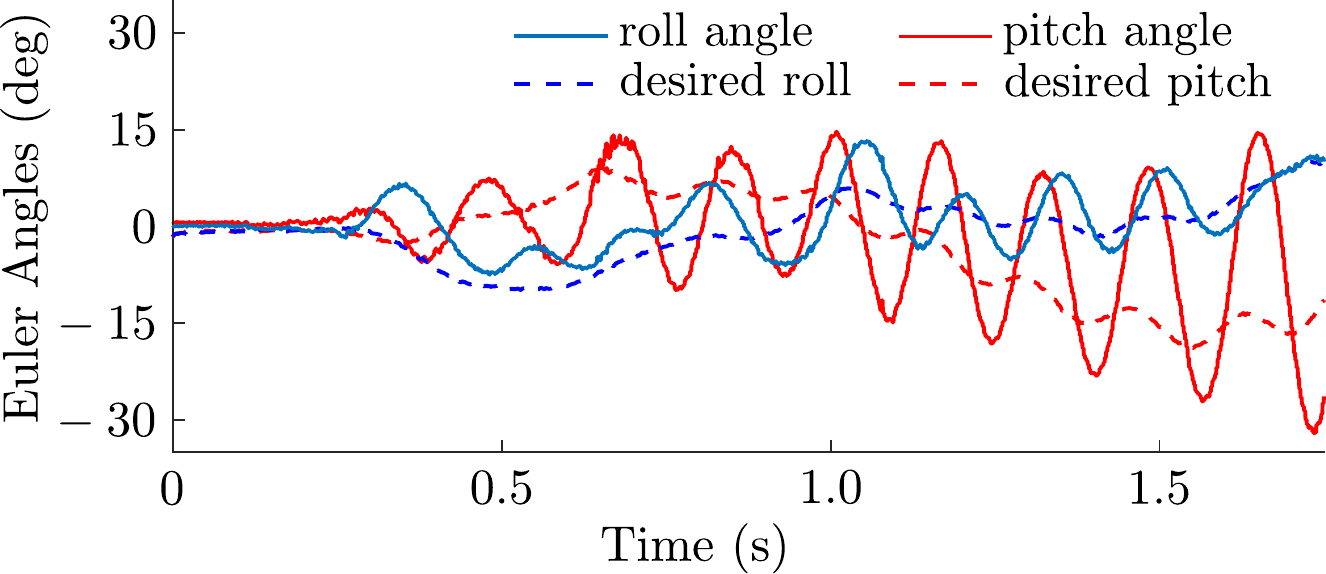}}
\caption{Position control experimental results. \textbf{A.}~The dash lines show the desired position of the center of mass and the solid lines show the measured position of the center of mass. \textbf{B.}~The dash lines show the desired Euler angles derived from (\ref{EQ13}) and (\ref{EQ14}); the solid lines show the measured Euler angles. \label{FigRAL06}}
\vspace{-2ex}
\end{figure}

\vspace{-0.5ex}
\label{SECTION05C}
In this experiment, a Bee\textsuperscript{+} prototype is commanded to hover at a desired position in space while driven by the attitude controller specified by (\ref{EQ10}) and the position controller described by (\ref{EQ11}) and (\ref{EQ12}). As in the experiment described in Section\,\ref{SECTION05B}, the desired and true yaw angles are assumed to be identical to each other, i.e. $\psi_{\ts{d}} = \psi$, which does not affect, in any way, the computation of position control signals as the direction of the total thrust force does not depend on the yaw angle. A photographic sequence of the experiment, with the corresponding time-lapse information, is shown in Fig.\,\ref{FigRAL05}-B; the associated experimental data is shown in Fig.\,\ref{FigRAL06}. Here, Fig.\,\ref{FigRAL06}-A shows the measured controlled position of the robot along with the corresponding reference signals. These data show that the robot approximately tracks the reference signals during the first second of the test; then, the position error along the $\bs{n}_1$ axis gradually increases. Fig.\,\ref{FigRAL06}-B shows that during the first second of the test, the measured roll and pitch angles approximately track the references and that the low-frequency content is tracked accurately; then, the pitch tracking error gradually becomes significant, which is consistent with the increasing position error along the $\bs{n}_1$ axis shown in Fig.\,\ref{FigRAL06}-A. We hypothesize that the oscillation about the pitch axis is caused by actuator saturation. This problem will be addressed by improving the robotic design in order to generate more thrust for position regulation and trajectory following. The complete set of experiments is presented in the supplementary movie S1.mp4.
\captionsetup[subfigure]{labelformat=empty}

\vspace{-0.5ex}
\section{Conclusions}
\vspace{-0.5ex}
\label{SECTION06}
We presented Bee\textsuperscript{+}, a 95-mg four-winged insect-sized flying robot with an extremely compact configuration. The key innovation that enabled the development of Bee\textsuperscript{+} is a new actuation technique based on the use of two pairs of twinned unimorph actuators. By employing instantaneous and time-averaged quasi-steady analyses and simulations, we estimated the main aerodynamic characteristics of the robotic design, including ranges for thrust forces, damping and steering torques. In addition, three different strategies for the generation of yaw torques were discussed and we determined that the ISP method is the most appropriated for the control of Bee\textsuperscript{+} prototypes. Finally, we presented a method for controller synthesis based on techniques developed for quadrotors and real-time control experiments. 
\ifCLASSOPTIONcaptionsoff
  \newpage
\fi

\bibliographystyle{IEEEtran}
\bibliography{paper}

\begin{thebibliography}{10}
\providecommand{\url}[1]{#1}
\csname url@samestyle\endcsname
\providecommand{\newblock}{\relax}
\providecommand{\bibinfo}[2]{#2}
\providecommand{\BIBentrySTDinterwordspacing}{\spaceskip=0pt\relax}
\providecommand{\BIBentryALTinterwordstretchfactor}{4}
\providecommand{\BIBentryALTinterwordspacing}{\spaceskip=\fontdimen2\font plus
\BIBentryALTinterwordstretchfactor\fontdimen3\font minus
  \fontdimen4\font\relax}
\providecommand{\BIBforeignlanguage}[2]{{%
\expandafter\ifx\csname l@#1\endcsname\relax
\typeout{** WARNING: IEEEtran.bst: No hyphenation pattern has been}%
\typeout{** loaded for the language `#1'. Using the pattern for}%
\typeout{** the default language instead.}%
\else
\language=\csname l@#1\endcsname
\fi
#2}}
\providecommand{\BIBdecl}{\relax}
\BIBdecl

\bibitem{Wood2008}
{R. J. Wood}, ``The first takeoff of a biologically inspired at-scale robotic
  insect,'' \emph{IEEE Trans. Robot.}, vol.~24, no.~2, pp. 341--347, 2008.

\bibitem{perez2011first}
N.~O. P{\'e}rez-Arancibia, K.~Y. Ma, K.~C. Galloway, J.~D. Greenberg, and R.~J.
  Wood, ``First controlled vertical flight of a biologically inspired
  microrobot,'' \emph{Bioinspir. Biomim.}, vol.~6, no.~3, p. 036009, 2011.

\bibitem{ma2013controlled}
K.~Y. Ma, P.~Chirarattananon, S.~B. Fuller, and R.~J. Wood, ``Controlled flight
  of a biologically inspired, insect-scale robot,'' \emph{Science}, 2013.

\bibitem{perez2015JINT}
{N. O. P{\'e}rez-Arancibia}, {P.-E. J. Duhamel}, {K. Y. Ma}, and {R. J. Wood},
  ``Model-free control of a hovering flapping-wing microrobot,'' \emph{J.
  Intell. Robot. Syst.}, vol.~77, no.~1, pp. 95--111, 2015.

\bibitem{graule2016perching}
M.~A. Graule, P.~Chirarattananon, S.~B. Fuller, N.~T. Jafferis, K.~Y. Ma,
  M.~Spenko, R.~Kornbluh, and R.~J. Wood, ``Perching and takeoff of a robotic
  insect on overhangs using switchable electrostatic adhesion,''
  \emph{Science}, vol. 352, no. 6288, pp. 978--982, 2016.

\bibitem{fuller2019four}
S.~B. Fuller, ``Four wings: An insect-sized aerial robot with steering ability
  and payload capacity for autonomy,'' \emph{IEEE Robot. Autom. Lett.}, vol.~4,
  no.~2, pp. 570--577, 2019.

\bibitem{dickinson1997function}
M.~H. Dickinson and M.~S. Tu, ``The function of dipteran flight muscle,''
  \emph{Comp. Biochem. Phys. A}, vol. 116, no.~3, pp. 223--238, 1997.

\bibitem{ma2012design}
K.~Y. Ma, S.~M. Felton, and R.~J. Wood, ``Design, fabrication, and modeling of
  the split actuator microrobotic bee,'' in \emph{Proc. 2012 IEEE Int. Conf.
  Intell. Robot. Syst.}, 2012, pp. 1133--1140.

\bibitem{doman2010wingbeat}
D.~B. Doman, M.~W. Oppenheimer, and D.~O. Sigthorsson, ``Wingbeat shape
  modulation for flapping-wing micro-air-vehicle control during hover,''
  \emph{J. Guid. Cont. Dynam.}, vol.~33, no.~3, pp. 724--739, 2010.

\bibitem{fuller2015rotating}
S.~B. Fuller, J.~P. Whitney, and R.~J. Wood, ``Rotating the heading angle of
  underactuated flapping-wing flyers by wriggle-steering,'' in \emph{Proc. 2015
  IEEE Int. Conf. Intell. Robot. Syst.}, 2015, pp. 1292--1299.

\bibitem{gravish2016anomalous}
N.~Gravish and R.~J. Wood, ``Anomalous yaw torque generation from passively
  pitching wings,'' in \emph{Proc. 2016 IEEE Int. Conf. Robot. Autom.}, 2016,
  pp. 3282--3287.

\bibitem{finio2012open}
B.~M. Finio and R.~J. Wood, ``Open-loop roll, pitch and yaw torques for a
  robotic bee,'' in \emph{Proc.2012 IEEE Int. Conf. Intell. Robot. Syst.},
  2012, pp. 113--119.

\bibitem{karasek2018tailless}
M.~Kar{\'a}sek, F.~T. Muijres, C.~D. Wagter, B.~D. Remes, and G.~C. de~Croon,
  ``A tailless aerial robotic flapper reveals that flies use torque coupling in
  rapid banked turns,'' \emph{Science}, 2018.

\bibitem{roshanbin2019yaw}
A.~Roshanbin and A.~Preumont, ``Yaw control torque generation for a hovering
  robotic hummingbird,'' \emph{Int. J. Adv. Robot. Syst.}, 2019.

\bibitem{jafferis2015design}
N.~T. Jafferis, M.~J. Smith, and R.~J. Wood, ``Design and manufacturing rules
  for maximizing the performance of polycrystalline piezoelectric bending
  actuators,'' \emph{Smart Mater. Struct.}, vol.~24, no.~6, p. 065023, 2015.

\bibitem{chang2018time}
L.~Chang and N.~O. P{\'e}rez-Arancibia, ``Time-averaged dynamic modeling of a
  flapping-wing micro air vehicle with passive rotation mechanisms,'' in
  \emph{AIAA Atmos. Flight Mech. Conf.}, 2018, p. 2830.

\bibitem{chang2016dynamics}
------, ``The dynamics of passive wing-pitching in hovering flight of flapping
  micro air vehicles using three-dimensional aerodynamic simulations,'' in
  \emph{AIAA Atmos. Flight Mech. Conf.}, 2016, p. 0013.

\bibitem{cheng2011translational}
B.~Cheng and X.~Deng, ``Translational and rotational damping of flapping flight
  and its dynamics and stability at hovering,'' \emph{IEEE Trans. Robot.},
  vol.~27, pp. 849--864, 2011.

\bibitem{Ying2018IROS}
Y.~Chen and N.~O. P\'erez-Arancibia, ``{Nonlinear adaptive control of quadrotor
  multi-flipping maneuvers in the presence of time-varying torque latency},''
  in \emph{{Proc. 2018 IEEE Int. Conf. Intell. Robot. Syst.}}, October 2018,
  pp. 1--9.

\bibitem{chen2017lyapunov}
Y.~Chen and N.~O. P{\'e}rez-Arancibia, ``Lyapunov-based controller synthesis
  and stability analysis for the execution of high-speed multi-flip quadrotor
  maneuvers,'' in \emph{2017 American Control Conference}.\hskip 1em plus 0.5em
  minus 0.4em\relax IEEE, 2017, pp. 3599--3606.

\bibitem{chirarattananon2014adaptive}
P.~Chirarattananon, K.~Y. Ma, and R.~J. Wood, ``Adaptive control of a
  millimeter-scale flapping-wing robot,'' \emph{Bioinspir. Biomim.}, vol.~9,
  no.~2, p. 025004, 2014.

\end{thebibliography}
\balance

\end{document}